\documentclass[10pt,twocolumn,twoside]{IEEEtran}
\usepackage{amsmath,graphicx}
\usepackage{amssymb}
\usepackage{graphicx}
\usepackage{graphics}
\usepackage{epstopdf}
\usepackage{amsmath}
\usepackage{algorithmicx}
\usepackage{algorithm}
\usepackage{algpseudocode}
\usepackage{multirow}
\usepackage{bm}
\usepackage{enumerate}
\usepackage{paralist}
\usepackage{array}

\usepackage[english]{babel}

\begin{document}
	
	\title{Learning Multiplication-free Linear Transformations}

	\author{Cristian Rusu\thanks{Cristian Rusu is with the
			Faculty of Automatic Control and Computer Science,
			University Politehnica of Bucharest, Romania
			(email: cristian.rusu@acse.pub.ro). This work was supported by
			the Romanian Ministry of Education and Research, CNCS-UEFISCDI,
			project number PN-III-P1-1.1-TE-2019-1843, within PNCDI III. Demo source code https://github.com/cristian-rusu-research/multiplication-free-transform
	
	Demo source code for \cite{FastSparsifyingTransforms} available at https://github.com/cristian-rusu-research/efficient-transform and for \cite{FastOvercomplete} available at https://github.com/cristian-rusu-research/efficient-overcomplete-transform}}
	
	\maketitle
	
	\begin{abstract}
		In this paper, we propose several dictionary learning algorithms for sparse representations that also impose specific structures on the learned dictionaries such that they are numerically efficient to use: reduced number of addition/multiplications and even avoiding multiplications altogether. We base our work on factorizations of the dictionary in highly structured basic building blocks (binary orthonormal, scaling and shear transformations) for which we can write closed-form solutions to the optimization problems that we consider. We show the effectiveness of our methods on image data where we can compare against well-known numerically efficient transforms such as the fast Fourier and the fast discrete cosine transforms.
	\end{abstract}

	\section{Introduction}
	In many situations, the success of theoretical concepts in signal processing applications depends on there existing an accompanying algorithmic implementation that is numerically efficient, e.g., Fourier analysis and the fast Fourier transform (FFT) or wavelet theory and the fast wavelet transform (FWT). Unfortunately, in a machine learning scenario where linear transformations are learned they do not exhibit in general advantageous numerical properties, as do the examples just mentioned, unless we explicitly search for such solutions.
	
	In this paper, we propose solutions to the dictionary learning problem \cite{DictMagazine} which construct linear transformations that have a series of desirable numerical properties while still sparsely representing the training data we supply. Our goal is to build these dictionaries $\mathbf{D}$ such that matrix-vector multiplications $\mathbf{Dx}$ have complexity $O(n \log n)$ or $O(n)$ while we also focus on investigating ways in which the number of multiplication operations can be reduced or completely avoided.
	
	There has been significant work in the literature to learn structured dictionaries that have controllable numerical complexity. One of the earlier attempts is to build a double sparse model \cite{DoubleSparsity} where the components of the dictionary are sparse linear combinations from a well-known transform that has a numerically efficient implementation. A recent paper \cite{FastOvercomplete} shows how to extend this model and also learn the numerically efficient transformation together with the sparse linear combinations. Other works focus on constructing dictionaries based on Kronecker products \cite{SeparableDictionary}, circulant (and union of circulants) \cite{BeforeCDLA, ShiftInvariantDictLearning} or convolutional \cite{ConvolutionalTree} structures, or square transformations that are factored by few Householder reflectors \cite{DictHouseholder}, Givens rotations \cite{lemagoarou:hal-01104696} and their generalization \cite{FastSparsifyingTransforms}.
	
	To our knowledge, the dictionary learning community has not investigated the possibility of constructing multiplication free linear transformations. This task has been well studied by the signal and image processing communities where integer-to-integer transformations (also called integer mappings) perform only addition and bit shift operations and therefore are essential for lossless compression. Fast multiplierless approximations of the discrete cosine transform based on a prototype method \cite{DyadicSymmetry}, a lattice structure \cite{MultiplierlessDCT}, the lifting scheme \cite{MultiplierlessDCT2}, an integer \cite{IntegerFFT} and an approximate multiplier-less \cite{1043869} fast Fourier transform were developed first. Then \cite{ReversibleIntegerMapping} introduced a general framework to build integer mappings from any linear transformation based on factorizations of (triangular and row) elementary reversible matrices and then showcases the framework on the discrete Fourier, cosine and wavelet transformations. Another general framework based on the general S transform is given in \cite{GST}. One image processing application is for the design of an integer color transform \cite{IntegerColorTransform}.
	
	In this paper, we will combine the benefits from both worlds: in the style of dictionary learning, we will learn a numerically efficient transformation from a training dataset that directly has an imposed structure to reduce or eliminate multiplication operations, in the fashion of integer mappings.
	
	The paper is structured as follows: in Section II we discuss ways to measure computational complexity, in Section III we briefly describe the dictionary learning problem and our computational design goals, then in Sections IV and V we develop the proposed learning procedures and finally in Section VI we show experimental results with image data where we compare to the discrete cosine transform.

	\section{A note on computational complexity}
	
	Given the number of computational platforms available today and their sophistication, our purpose is not to provide an exhaustive, detailed discussion of the subject but to give arguments that multiplication-free algorithms are relevant.
	
	In most scenarios the computational complexity accounts for all the operations performed by the system, i.e., we count together mathematical operations like additions, multiplications etc. When considering modern computing systems this choice is a natural one: these mathematical operations take approximately the same time sophisticated hardware. For example, numerical simulation performed with an Intel i$7^\text{\tiny{\textcopyright}}$ processor shows that integer multiplication is on average approximately only 10\% slower than integer addition (running Linux, using the gcc with the --O3 flag, the program performs the operations on random integer operands in two arrays and the results are stored in a third). Historically, this was not the case. In the past, computer scientists have made several efforts to reduce the number of multiplications in their algorithmic implementations in favor of performing more addition operations. A classic example is the multiplication of two complex numbers 
	which can be done in two ways:
	the first takes four multiplications and two additions while the second one has three multiplications and five additions, i.e., it was computationally convenient to replace one multiplication by three addition operations. Even so, modern computing systems still perform integer addition faster than integer multiplication in general (one clock cycle versus three to ten clock cycles depending on the particular processor)\footnote{Intel 64 and {IA}-32 Architectures Optimization Reference Manual}.
	
	From an algorithmic perspective, for numbers represented using $n$ bits, it is well understood that integer addition has complexity $O(n)$ while the best asymptotic bound $O(n \log n \log \log n)$ for integer multiplication, given for realistically reasonably large $n$, is achieved by the Schonhage-Strassen algorithm \cite{Strassen1971}. Other, asymptotically less efficient approaches, include Karatsuba's algorithm \cite{Karatsuba} and, its generalization, the Toom-Cook algorithm \cite[Section 9.5]{TC} -- both use techniques similar to the previously described trick of replacing an intermediate multiplication operation with several additions achieve complexity $n^{\log 3}$ and $O\left(n^{\frac{\log 5}{\log 3}}\right)$, respectively. In terms of hardware, the modern multiplier architectures use the Baugh-Wooley algorithm \cite{BaughWooley} or Wallace trees \cite{WT}, for example. These methods reduce the performance gap with binary addition (as previously observed experimentally) at the cost of increasing the complexity of the circuitry.
	
	Aside from the execution time, there are several other important complexity measures, like power consumption and circuitry size, especially when considering some custom or embedded computational platforms where low size, weight, power and cost (SWaP-C) solutions are preferred. An $n$ bit full-adder needs $5n$ logic gates: one OR, two AND and two XOR gates per bit. In the case of binary multiplication, for example, the relatively simple sequential $n$ bit array multiplier needs $31n$ gates: the $n$ and $2n$ bit registers consist of $15n$ gates, the ALU contains an adder and two multiplexers which consist of $16n$ gates ($5n$ gates for the adder, $4n$ gates for the $2\times 1$ mux and $7n$ gates for the $4 \times 1$ mux).
	
	Application-specific integrated circuits (ASICS) are circuits that are designed to perform only one (or a small set) of tasks, unlike CPUs. Field-programmable gate arrays (FPGAs) are a computational platform that belongs to the ASICS class. Addition operations in FPGAs are generally performed using look-up tables while for the multiplications some specialized extra components are needed (like DSP slices). Here, a frequently used performance indicator is the power-delay product (the product between the energy consumption and the input-output delay of a circuit). For a very popular FPGA computer-aided design tools for arithmetic code generation, the $\text{Xilinx}^\text{\tiny{\textcopyright}}$ IP Core Generator, with 32-bit operands the addition operation has a power-delay product of 0.67 nJ (see Table 4.1 of \cite{HighPerformanceFPGA}) while the 15-bit multiplication operation has a power-delay product of 2.21 nJ (see Table 4.3 of \cite{HighPerformanceFPGA}).
	
	The same power-delay product values for microprocessors at 45nm are given in \cite{Horowitz2014} to be: 0.1pJ and 3pJ for 32-bit integer addition and multiplication, respectively, and 0.9pJ and 4pJ for 32-bit floating point addition and multiplication, respectively. These numbers do not take into account memory access latency and power consumption (which according to \cite{Horowitz2014} are also a major contributor to the overall power consumption).
	
	\section{Preliminaries}
	
	In this section, we briefly describe the general dictionary learning problem and then proceed to list some computationally-desirable properties of the learned dictionary. We also describe some basic matrix building blocks that we will use in this paper to reach the desirable properties listed.

	\subsection{The dictionary learning problem}
	
	Given a $N$ point dataset $\mathbf{Y} \in \mathbb{R}^{n \times N}$ and the average sparsity $s \in \{1, \dots, n-1\}$, the dictionary learning problem can be stated in the optimization language as
	\begin{equation}
	\begin{aligned}
	\underset{\mathbf{D},\ \mathbf{X}}{\text{minimize}} & & & \| \mathbf{Y} - \mathbf{DX}  \|_F^2 \\
	\text{subject to} & & & \text{diag}(\mathbf{D}^T \mathbf{D}) = \mathbf{1}_{n \times 1},\\
	& & &  \| \text{vec}(\mathbf{X}) \|_0 \leq sN,
	\end{aligned}
	\label{eq:dlaproblem}
	\end{equation}
	where $\mathbf{D} \in \mathbb{R}^{n \times n}$, which always has unit $\ell_2$ columns, is called the dictionary (in general $\mathbf{D}$ can be overcomplete, but in this paper we consider square dictionaries which we also call transforms \cite{TransLearning2013}) and $\mathbf{X} \in \mathbb{R}^{n \times N}$ has the sparse representations of all data points in the dictionary $\mathbf{D}$. The $\ell_0$ pseudo-norm constraint, which counts the number of non-zero in the matrix $\mathbf{X}$, ensures that on average each data point is represented using $s$ columns (also called atoms) from $\mathbf{D}$. The dictionary learning problem is hard in general, so most optimization techniques lead to local minima points of \eqref{eq:dlaproblem} by a process of alternating minimization: keep $\mathbf{D}$ fixed and create $\mathbf{X}$ and then vice-versa. In this paper, we deploy the same alternating technique and we focus on constructing the dictionary $\mathbf{D}$ such that it has some specific computational properties. We do not focus on how to construct the sparse representations $\mathbf{X}$, i.e., we will use the appropriate, well-established algorithms from the literature to build $\mathbf{X}$ \cite[Chapter 1]{DumitrescuBook2018}. 
	
	For a recent, detailed description of the dictionary learning problem and some of its solutions the reader is encouraged to check \cite[Chapters 2 and 3]{DumitrescuBook2018}.
	
	\subsection{The basic building block}
	
	Based on the work in \cite{FastSparsifyingTransforms}, we revise the $n \times n$ R-transform:
	\begin{equation}
	\mathbf{R}_{ij} = \begin{bmatrix} 
	\mathbf{I}_{i-1} &  &  &  & \\
	& a &  & c & \\
	& & \mathbf{I}_{j-i-1} & & \\
	& b & & d & \\
	& & & & \mathbf{I}_{n-j} \\
	\end{bmatrix}, \mathbf{\tilde{R}} = \begin{bmatrix}
	a & c \\
	b & d
	\end{bmatrix}.
	\label{eq:Rtransform}
	\end{equation}
	These matrices can be viewed as perturbations of the identity matrix: $\mathbf{R}_{ij}$ has zero entries everywhere except for its diagonal (with entries $a$ and $d$ in positions $(i,i)$ and $(j,j)$, respectively, and the rest with value one) and the only two off-diagonal entries $c$ and $b$, on positions $(i,j)$ and $(j,i)$ respectively. The subscripts of the R-transform define the rows on which the non-trivial values are stored. As it is convenient in many situations to reference the unique part of $\mathbf{R}_{ij}$ separately we denote it by $\mathbf{\tilde{R}}$. Through this paper we will use the same template for the transformations we propose: $n \times n$ identity except for 2 coordinates where we will perform a carefully chosen $2 \times 2$ calculation. We define transformations that are products of $m$ basic building blocks, like
	\begin{equation}
	\mathbf{R} = \prod_{k=1}^m \mathbf{R}_{i_k j_k} = \mathbf{R}_{i_m j_m} \dots \mathbf{R}_{i_1 j_1}.
	\end{equation}
	Analogously to \eqref{eq:Rtransform}, we denote the unique $2 \times 2$ part of each $\mathbf{R}_{i_k j_k}$ by $\mathbf{\tilde{R}}_k$.
	
	Given $\mathbf{x} \in \mathbb{R}^{n}$ the matrix-vector multiplication $\mathbf{R}_{i_k j_k} \mathbf{x}$ takes four multiplications and two addtions. To avoid these multiplications, similarly to fixed point number representations, we use the sums of powers of two (SOPOT) set:
	\begin{equation}
	\mathcal{R}_p = \left\{ x \ \Big| \ x = \sum_{t = 1}^{p} s_t 2^{v_t} ; \ s_t \in \{\pm 1 \},\ v_t \in \mathbb{Z} \right\},
	\label{eq:thesetP}
	\end{equation}
	where the parameter $p$ establishes the precision of the entries. By convention when $p = \infty$ we use the precision of the working data type (double floating point in our case). For given $p$, we provide in Algorithm 1 an iterative greedy procedure to compute the representation of any real scalar input $x$ in the set $\mathcal{R}_p$. We use this set to represent our transformations $\mathbf{\tilde{R}}$ which we now denote $\mathbf{\tilde{R}}_{k, p} \in \mathcal{R}_p^{2 \times 2}$, i.e., we approximately represent each $a,b,c,d$ in the set $\mathcal{R}_p$, and we call the overall $\mathbf{R}_{i_k j_k, p}$ an R$_p$-transform.
	\begin{algorithm}[t]
		\caption{ \textbf{-- Representation in $\mathcal{R}_p$. } \newline \textbf{Input: } The real value $x \in \mathbb{R}$ and the precision $p \in \mathbb{N}^{*}$. \newline \textbf{Output: } The value $y \in \mathcal{R}_p$ and its representation, i.e., $\mathbf{s} \in \{\pm 1 \}^p$ and $\mathbf{v} \in \mathbb{Z}^p$, closest to $x$ in absolute value.}
		\begin{algorithmic}
			\State \textbf{1. } Initialize residual $r = x$ and current estimate $y = 0$.
			
			\State \textbf{2. } For $i=1,\dots,p:$
			\begin{itemize}
				\item Set $s_i = \text{sign}(r)$ and $v_i = \underset{k \in \mathbb{Z}}{\arg \min} ||r| - 2^k|$.
				
				\item Update estimate $y = y + s_i 2^{v_i}$ and residual $r = x - y$.
			\end{itemize}
		\end{algorithmic}
	\end{algorithm}
	
	Structures like these R$_p$-transforms are interesting numerically because matrix-vector multiplication takes $4p$ bit shifts and $4p-2$ additions ($4p-4$ to form the four products and 2 to add the results for line $i$ and $j$ respectively). This is because scalar multiplication with $a \in \mathcal{R}_p$ takes $p$ bit shifts and $p-1$ additions. In terms of the coding complexity, assuming $8$ bits are used to store each $v_t$ in \eqref{eq:thesetP}, storing $\mathbf{R}_{i_k j_k,p}$ takes approximately $36p + 2\log_2 n$ bits ($2 \log_2 n$ bits to store the indices $i,j$ and $9p$ bits to store each entry in $\mathcal{R}_p$).
	
	In this paper, we propose to learn dictionaries which are products of basic transformations like \eqref{eq:Rtransform}, while we also impose some additional constraints, e.g., orthogonality or some specific arithmetic structure for the non-zero entries.
	
	\subsection{The computational properties for the dictionary}
	
	In this section we define and present properties of the basic building blocks we consider for numerically efficient factorizations that will allow us to achieve our design goals. 
	
	Our goal is to construct a dictionary $\mathbf{D} \in \mathbb{R}^{n \times n}$ that display properties such as:
	\begin{enumerate}
		\item[P1.] The computational complexity (the number of additions, bit shifts and multiplications) of $\mathbf{Dx}$ and $\mathbf{D}^{-1} \mathbf{x}$ for any given $\mathbf{x} \in \mathbb{R}^n$ is controllable, preferably $O(n \log n)$;
		
		\item[P2.] If $\mathbf{x} \in \mathcal{R}_{p}^{n}$ then $\mathbf{Dx} \in \mathcal{R}_{p \prime}^{n}$, i.e., if $\mathbf{x}$ has a fixed point representation then so does $\mathbf{Dx}$; if $\mathbf{Dx}$ is calculated exactly in $\mathcal{R}_{p}^{n}$ then $\mathbf{D}^{-1} \mathbf{x}$ can also be calculated exactly in $\mathcal{R}_{p \prime}^{n}$ with $p \prime \neq p$, i.e., if $\mathbf{D}$ has a fixed representation then so does the inverse $\mathbf{D}^{-1}$;
		
		\item[P3.] $\mathbf{D}$ is exactly reversible, i.e., $\mathbf{D}^{-1} \mathbf{D} = \mathbf{I}$, when $\mathbf{D}$ has a fixed point representation;
		
		\item [P4.] Reduce, and ideally, eliminate multiplication operations for $\mathbf{Dx}$ and $\mathbf{D}^{-1}\mathbf{x}$;
		
	\end{enumerate}
	
	In the following sections, we will distinguish between orthonormal (Section IV) and general (Section V) dictionary learning procedures and discuss how the transformations we learn achieve some or all of these desirable properties.
	
	\section{The orthonormal case}
	
	In this section, we propose two orthonormal dictionary learning algorithms: one with a reduced number of multiplications and one that avoids completely such operations.
	
	\subsection{Numerically efficient orthogonal transforms: B$_m$--DLA}
	
	We define the two sets of orthonormal binary $2 \times 2$ matrices:
	\begin{equation}
	\begin{aligned}
	\mathcal{G}_1 \! = \! \frac{1}{\sqrt{2}} \! \left\{ \begin{bmatrix}
	-1 & 1 \\
	1 & 1
	\end{bmatrix},\begin{bmatrix}
	1 & 1 \\
	-1 & 1
	\end{bmatrix},\begin{bmatrix}
	1 & -1 \\
	1 & 1
	\end{bmatrix},\begin{bmatrix}
	1 & 1 \\
	1 & -1
	\end{bmatrix},  \right. \\
	\left. \begin{bmatrix}
	1 & -1 \\
	-1 & -1
	\end{bmatrix},\begin{bmatrix}
	-1 & -1 \\
	1 & -1
	\end{bmatrix},\begin{bmatrix}
	-1 & 1 \\
	-1 & -1
	\end{bmatrix},\begin{bmatrix}
	-1 & -1 \\
	-1 & 1
	\end{bmatrix} \right\},
	\end{aligned}
	\label{eq:localstructure1}
	\end{equation}
	\begin{equation}
	\begin{aligned}
	\text{and } \mathcal{G}_2 \! = \! \left\{ \begin{bmatrix}
	0 & 1  \\
	-1 & 0
	\end{bmatrix},\begin{bmatrix}
	0 & -1 \\
	1 & 0
	\end{bmatrix},\begin{bmatrix}
	1 & 0 \\
	0 & -1
	\end{bmatrix},\begin{bmatrix}
	-1 & 0 \\
	0 & 1
	\end{bmatrix}, \right. \\
	\left. \begin{bmatrix}
	0 & -1  \\
	-1 & 0
	\end{bmatrix},\begin{bmatrix}
	-1 & 0 \\
	0 & -1
	\end{bmatrix},\begin{bmatrix}
	0 & 1 \\
	1 & 0
	\end{bmatrix}, \begin{bmatrix}
	1 & 0 \\
	0 & 1
	\end{bmatrix} \right\}.
	\end{aligned}
	\label{eq:localstructure1b}
	\end{equation}
	
	We define an orthogonal B-transform denoted $\mathbf{B}_{ij}$ as a constraint R-transform in \eqref{eq:Rtransform} where we have the non-trivial, non-zero $2 \times 2$ part positioned at indices $i$ and $j$, which we denote $\mathbf{\tilde{B}}_{ij}$, defined as one of the sixteen options, i.e., $\mathbf{\tilde{B}}_{ij} \in \{ \mathcal{G}_1 \cup \mathcal{G}_2\}$. Notice that given $\mathbf{A}, \mathbf{B} \in \mathcal{G}_2$ we have that $\mathbf{AB} \in \mathcal{G}_2$, $\mathbf{BA} \in \mathcal{G}_2$ (and in fact $\mathcal{G}_1 \cup \mathcal{G}_2$ has a group structure) and given $\mathbf{C} \in \mathcal{G}_1$ we have that $\mathbf{AC} \in \mathcal{G}_1$ and $\mathbf{CA} \in \mathcal{G}_1$. The factor $2^{-\frac{1}{2}}$ is there to keep the B-transforms orthonormal (orthogonal and with columns normalized in $\ell_2$).
	
	Structures like B-transforms are useful because matrix-vector multiplication takes four operations: two additions and two multiplications (both by $2^{-\frac{1}{2}}$). The coding complexity of storing $\mathbf{B}_{ij} \in \mathbb{R}^{n \times n}$ is approximately $4 + 2\log_2 n$ bits (the first term encodes the choice in $\mathcal{G}_1 \cup \mathcal{G}_2$ while the second encodes the two indices $i$ and $j$). We could avoid the multiplications by approximating $2^{-\frac{1}{2}}$ in $\mathcal{R}_p$ but we would lose orthogonality.
	
	We are interested in solving optimization problems that consider one B-transform as a dictionary:
	\begin{equation}
	\| \mathbf{Y} - \mathbf{B}_{ij} \mathbf{X} \|_F^2 = \| \mathbf{Y} \|_F^2 + \| \mathbf{X} \|_F^2 - 2\text{tr}(\mathbf{Z}) + C_{ij}^{(t)},
	\label{eq:onlyoneB}
	\end{equation}
	where we have used the definition of the Frobenius norm $\| \mathbf{A} \|_F^2 = \text{tr}(\mathbf{A}^T \mathbf{A})$ and the index $t=1,\dots,15,$ runs through the possible variants $\mathbf{\tilde{B}}_{ij}$ in \eqref{eq:localstructure1}, \eqref{eq:localstructure1b}. Therefore, we have for the order of the transformations in \eqref{eq:localstructure1}, \eqref{eq:localstructure1b}:
	\begin{equation}
	\begin{aligned}
	C_{ij}^{(1)} \! \! = &  c_1 Z_{ii}\! + \! c_2 Z_{jj} \! -\! \sqrt{2}Z_{so}, \ C_{ij}^{(2)} \!\! = \!c_2Z_{sd} \!-\! \sqrt{2}Z_{do}, \\
	C_{ij}^{(3)} \! \! = &  c_2Z_{sd}\! +\! \sqrt{2}Z_{do},\ C_{ij}^{(4)} \!\! = \!c_2 Z_{ii} \! + \! c_1 Z_{jj} \!-\! \sqrt{2}Z_{so}, \\
	C_{ij}^{(5)} \! \! = & c_2 Z_{ii}\! + \! c_1 Z_{jj} \! +\! \sqrt{2}Z_{so},\ C_{ij}^{(6)} \!\! = \!c_1 Z_{sd} \!+\! \sqrt{2}Z_{do},\\
	C_{ij}^{(7)} \! \! = & c_1Z_{sd}\! -\! \sqrt{2}Z_{do},\ C_{ij}^{(8)} \!\!=\! c_1 Z_{ii} \!+\! c_2 Z_{jj} \!+\! \sqrt{2}Z_{so}, \\
	C_{ij}^{(9)} = & 2(Z_{sd} - Z_{do}),\ C_{ij}^{(10)} = 2(Z_{sd} + Z_{do}), \\
	C_{ij}^{(11)} = & 4Z_{jj},\ C_{ij}^{(12)} = 4Z_{ii},\ C_{ij}^{(13)} = 2(Z_{sd} + Z_{so}), \\
	C_{ij}^{(14)} = & 4Z_{sd},\ C_{ij}^{(15)} = 2(Z_{sd} - Z_{so}),\ C_{ij}^{(16)} = 0,
	\end{aligned}
	\label{eq:1to4}
	\end{equation}
	with $Z_{sd} = Z_{ii} + Z_{jj}, Z_{so} = Z_{ij} + Z_{ji}, Z_{do} = Z_{ij} - Z_{ji}$, the constants $c_1 = 2 + \sqrt{2},\ c_2 = 2 - \sqrt{2}$ and we define:
	\begin{equation}
	\mathbf{Z} = \mathbf{YX}^T \text{ with entries } Z_{ij} = \mathbf{y}_i^T \mathbf{x}_j,
	\label{eq:theZ}
	\end{equation}
	where $\mathbf{y}_i^T$ and $\mathbf{x}_i^T$ are the $i^\text{th}$ rows of $\mathbf{Y}$ and $\mathbf{X}$, respectively.
	
	The $\mathbf{B}_{i j}$ that minimizes \eqref{eq:onlyoneB} is given by
	\begin{equation}
	(i^\star, j^\star, t^\star) = \underset{t,i<j}{\arg \min} \ C_{i j}^{(t)},
	\label{eq:Bargmin}
	\end{equation}
	for the $C_{i j}^{(t)}$ in \eqref{eq:1to4} with $t = 1,\dots,15,$ and $j=1,\dots,n-1$. The total computational complexity to find the minimizer of \eqref{eq:Bargmin} is: $2n^2N$ operations to construct $\mathbf{Z}$, which dominates the computational complexity; $O(n^2)$ operations to compute all the fifteen $C_{ij}^{(t)}$ for all $\frac{n(n-1)}{2}$ distinct pairs $(i,j)$ with $i < j$ and solve \eqref{eq:Bargmin}; and $2n$ operations to compute $2\text{tr}(\mathbf{Z})$.
	\begin{algorithm}[t]
		\caption{ \textbf{-- B$_m$--DLA. } \newline \textbf{Input: } The dataset $\mathbf{Y} \in \mathbb{R}^{n \times N}$, the sparsity $s$ and the number of B-transforms $m$ in the dictionary.\newline \textbf{Output: } The orthonormal transformation $\mathbf{B}$ \eqref{eq:theBtransform} composed of $m$ B-transforms and the sparse representations $\mathbf{X}$ such that $\| \mathbf{Y} - \mathbf{BX} \|_F^2$ is reduced.}
		\begin{algorithmic}
			\State \textbf{1. } Initialize transform: set $\mathbf{B}_{i_k j_k} = \mathbf{I}_{ n \times n}$ for $k=1,\dots,m$.
			
			\State \textbf{2. } Initialize sparse representations: compute the singular value decomposition of the dataset $\mathbf{Y} = \mathbf{U \Sigma V}^T$ and compute the sparse representations $\mathbf{X} = \mathcal{T}_s (\mathbf{U}^T \mathbf{Y})$.
			
			\State \textbf{3. } For $1,\dots,K:$
			\begin{itemize}
				\item Compute $\mathbf{Z} = \mathbf{YX}^T$ and all scores $C_{ij}^{(t)}$ from \eqref{eq:1to4} for $i=1,\dots,n-1,\ j = i+1,\dots,n$ and $t=1,\dots,15$.
				\item For $k=1,\dots,m$ update all $\mathbf{B}_{i_k j_k}$, for each $k$:
				\begin{itemize}
					\item With all $\mathbf{B}_{i_q j_q},\ q \neq k,$ fixed, compute the new $\mathbf{B}_{i_k j_k}$ the minimizer of \eqref{eq:theBobjectivefunction} by \eqref{eq:Bargmin} with $\mathbf{Z}_k = \mathbf{Y}_k\mathbf{X}_k^T$.
					\item Update scores $C_{i j_k}^{(t)}$ and $C_{i_k j}^{(t)}$ for $t=1,\dots,15$.
				\end{itemize}
				\item Compute new sparse representations $\mathbf{X} = \mathcal{T}_s (\mathbf{B}^T \mathbf{Y})$.
			\end{itemize}
		\end{algorithmic}
	\end{algorithm}
	
	We propose a method to learn orthogonal dictionaries that are factorized as a product of $m$ B-transforms. Therefore we propose the following structure for our learned dictionary:
	\begin{equation}
	\mathbf{B} = \prod_{k=1}^{m} \mathbf{B}_{i_k j_k} = \mathbf{B}_{i_m j_m} \dots \mathbf{B}_{i_1 j_1}.
	\label{eq:theBtransform}
	\end{equation}
	With this choice, the dictionary learning objective function for a single transformation indexed $k$ is:
	\begin{equation}
	\| \mathbf{Y} -  \mathbf{B} \mathbf{X} \|_F^2
	= \| \mathbf{Y}_k - \mathbf{B}_{i_k j_k} \mathbf{X}_k \|_F^2,
	\label{eq:theBobjectivefunction}
	\end{equation}
	where $\mathbf{Y}_k = \prod_{q=k+1}^{m} \mathbf{B}_{i_q j_q}^T \mathbf{Y}$ and $\mathbf{X}_k = \prod_{q=1}^{k-1} \mathbf{B}_{i_q j_q} \mathbf{X}$. In this development we have used the fact that orthonormal transformations are invariant in the Frobenius norm, i.e., $\| \mathbf{QY} \|_F = \| \mathbf{Q}^T \mathbf{Y} \|_F = \| \mathbf{Y} \|_F$ for any orthonormal $\mathbf{Q} \in \mathbb{R}^{n \times n}$. Notice that we have reduced the objective function to the form in \eqref{eq:onlyoneB}. Therefore, we propose an efficient iterative process that updates a single $\mathbf{B}_{i_k j_k}$ at a time while keeping the others fixed until all components are optimized.
	
	We describe the full learning procedure in Algorithm 2. This algorithm updates iteratively each B-transform in the composition of the dictionary $\mathbf{B}$ and then the sparse representations $\mathbf{X}$. Since each step is solved exactly to optimality, the algorithm monotonically converges overall to a local minimum point or stops early in the maximum $K$ iterative steps. In the description of the algorithms, we have used $\mathcal{T}_s( )$, an operator that given a matrix sets to zero, in each column separately, all entries except the highest $s$ in magnitude.
	
	\noindent \textbf{Remark 1 (Working with limited precision).} In order to design exactly invertible orthogonal linear transformations when using data with fixed bit representation, lifting schemes \cite{Lifting1, Lifting2} were introduced in the past: $\begin{bmatrix}
	c & -s \\
	s & c
	\end{bmatrix} =  \begin{bmatrix}
	1 & \frac{c-1}{s} \\
	0 & 1
	\end{bmatrix} \begin{bmatrix}
	1 & 0\\
	s & 1
	\end{bmatrix} \begin{bmatrix}
	1 & \frac{c-1}{s} \\
	0 & 1
	\end{bmatrix}, 
	\begin{bmatrix}
	c & s \\
	s & -c
	\end{bmatrix} =  \begin{bmatrix}
	1 & \frac{c-1}{s} \\
	0 & 1
	\end{bmatrix} \begin{bmatrix}
	1 & 0\\
	s & 1
	\end{bmatrix} \begin{bmatrix}
	1 & -\frac{c-1}{s} \\
	0 & -1
	\end{bmatrix}$.
	The proposed B-transforms can naturally be implemented with these schemes, as they are particular $2 \times 2$ orthonormal matrices. For the transformations in $\mathcal{G}_1$ we have that $\frac{c-1}{s} \in \{ \pm (1-\sqrt{2}), \pm (1+\sqrt{2})  \}$ and $s \in \left\{ \pm 2^{-\frac{1}{2}} \right\}$ while the transformations in $\mathcal{G}_2$ are multiplier-less and therefore do not need the lifting scheme representations. Still, note that matrix-vector multiplications with matrices from $\mathcal{G}_1$ take two multiplications and two additions, while using lifting schemes representations needs three multiplications and three additions. The elements $\frac{c-1}{s}$ and $s$ in the lifting scheme can be represented in $\mathcal{R}_p$ to avoid multiplications altogether, but at the cost of loosing orthogonality.$\hfill \blacksquare$

	\noindent \textbf{Remark 2 (Avoiding the normalization factor).} The normalization by $2^{-\frac{1}{2}}$ seems to cause complications from a numerically efficiency stand point. Therefore, we could define an O-transform $\mathbf{O}_{ij} \in \mathbb{R}^{n \times n}$ which is achieved for \eqref{eq:Rtransform} when $a,b,c,d \in \{ \pm 1 \}$ such that $\mathbf{\tilde{R}}_{ij}$ is orthogonal. These eight transformations use $\sqrt{2} \mathbf{\tilde{B}}_{ij}$ for the structure \eqref{eq:localstructure1}. Solving such least squares problems leads to
	\begin{equation}
	\| \mathbf{Y} - \mathbf{O}_{ij} \mathbf{X} \|_F^2 = \| \mathbf{Y} \|_F^2 + \| \mathbf{X} \|_F^2 - 2\text{tr}(\mathbf{Z}) + H_{ij}^{(t)},
	\label{eq:onlyoneQ}
	\end{equation}
	where, for $t=1,\dots,8$, we have defined
	$H_{ij}^{(t)} =  W_{ii} + W_{jj} - 2(a-1)Z_{ii} - 2(d-1)Z_{jj} - 2bZ_{ij}-2cZ_{ji}+2(ab+cd)W_{ij}$.
	We have used the notation:
	\begin{equation}
	\mathbf{W} = \mathbf{X}\mathbf{X}^T\text{ with entries } W_{ij} = \mathbf{x}_i^T \mathbf{x}_j.
	\label{eq:theW}
	\end{equation}
	where $\mathbf{x}_i^T$ is the $i^\text{th}$ row of $\mathbf{X}$.
	
	Similarly to \eqref{eq:theBtransform}, based on these simple transformation we can define $\mathbf{O} = \prod_{k=1}^{m} \mathbf{O}_{i_k j_k} = \mathbf{O}_{i_m j_m} \dots \mathbf{O}_{i_1 j_1}.$
	Because the normalization is entirely avoided, $\mathbf{O}_{ij}$ is no longer orthogonal and therefore $\mathbf{O}$ is not orthogonal. A drawback of this is the fact that the transformations $\mathbf{O}_{i_k j_k}$ cannot be rearranged as in \eqref{eq:theBobjectivefunction}, making the update of an individual transformation while keeping all others fixed more difficult.
	
	But this structure has the advantage of completely avoiding multiplication operations, i.e., matrix-vector multiplications $\mathbf{O}_{ij} \mathbf{x}$ take 2 addition operations. Also, notice that $\mathbf{O}$ has integer entries and that $\det (\mathbf{O}) = \pm 2^m$ since $\det (\mathbf{O}_{ij}) = \pm 2$.
	
	A sufficient condition for the local optimality of $\mathbf{O}_{ij}$ is that $\| \mathbf{x}_j \|_2 - \| \mathbf{y}_i - \mathbf{x}_i \|_2 \geq 0, \ \forall\ i \neq j$, i.e., the energy of any row error is less than the energy of all other rows of $\mathbf{X}$.
	
	\noindent \textbf{Proof.} Given any rows of $\mathbf{Y}$ and $\mathbf{X}$ there is no O-transform that improves the objective function if $\| \mathbf{y}_i - \mathbf{x}_i \|_2^2 \leq \| \mathbf{y}_i - (\mathbf{x}_i \pm \mathbf{x}_j) \|_2^2,\ \forall \ i \neq j$. Developing this leads to $\| \mathbf{x}_j \|_2^2 - |C| \geq 0,\ C = 2\mathbf{x}_i^T\mathbf{x}_j-2\mathbf{y}_i^T \mathbf{x}_j$.
	By the Cauchy-Schwartz inequality we have that $-|C| \geq -\| \mathbf{x}_j \|_2\| \mathbf{y}_i - \mathbf{x}_i \|_2$ and therefore $\| \mathbf{x}_j\|_2^2 - |C| \geq \| \mathbf{x}_j\|_2^2 - \| \mathbf{x}_j \|_2 \| \mathbf{y}_i - \mathbf{x}_i \|_2 \geq 0$. $\hfill \blacksquare$
	
	The matrix-vector multiplication operation with the whole $\mathbf{B}$ transformation takes $2m$ additions and $2m$ multiplications with the same constant value $2^{-\frac{1}{2}}$.

	Related to our previously stated desired computational properties we have that: P1 is achieved by taking the number of B-transforms in $\mathbf{B}$ to be $m \sim O(n \log n)$ and since $\mathbf{B}^{-1} = \mathbf{B}^T$ we have that the inverse transformation enjoys the same computational complexity (the inverse of any B-transforms is itself a B-transform); P2 can be achieved by using the lifting schemes and representing $\frac{c-1}{s}$ and $s$ in $\mathcal{R}_p$ but notice that with the fixed point representations we no longer have an orthogonal transformation, we denote $\mathbf{B}_p$ the transformation $\mathbf{B}$ in the lifting scheme with elements in $\mathcal{R}_p$; using the representation explained for P2 we also cover the requirement P3; regarding P4, notice that multiplication with each B-transform takes two multiplications as compared to three in the general lifting scheme.
	
	\subsection{Multiplication-free orthogonal transforms: M$_q$--DLA}
	\begin{algorithm}[t]
		\caption{ \textbf{-- M$_q$--DLA. } \newline \textbf{Input: } The dataset $\mathbf{Y} \in \mathbb{R}^{n \times N}$, the sparsity $s$ and the number of stages $q$ in the dictionary.\newline \textbf{Output: } The orthonormal transformation $\mathbf{M}$ \eqref{eq:theStransforms} composed of $q$ stages of $\frac{n}{2}$ B-transforms and the sparse representations $\mathbf{X}$ such that $\| \mathbf{Y} - \mathbf{MX} \|_F^2$ is reduced.}
		\begin{algorithmic}
			\State \textbf{1. } Initialize transform to the identity matrix, set $\mathbf{M} = \mathbf{I}$ by $\mathbf{M}_\ell = \mathbf{I}$ for $\ell = 1,\dots,q$.
			
			\State \textbf{2. } Initialize sparse representations: compute the singular value decomposition of the dataset $\mathbf{Y} = \mathbf{U \Sigma V}^T$ and compute the sparse representations $\mathbf{X} = \mathcal{T}_s (\mathbf{U}^T \mathbf{Y})$.
			
			\State \textbf{3. } For $\ell = 1,\dots,q:$
			\begin{itemize}
				\item Compute $\mathbf{Z}_\ell = 2^{-\frac{\ell-1}{2}}\mathbf{Y} (\mathbf{M}_{\ell-1} \cdots \mathbf{M}_1 \mathbf{X})^T$.
				
				\item Using $\mathbf{Z}_\ell$, compute all scores $C_{ij}^{(t)}$ from \eqref{eq:1to4} for $i=1,\dots,n-1,\ j = i+1,\dots,n$ and $t=1,\dots,8$.
				
				\item  Compute $C_{i_k j_k}^{(0)}$ and update all $\mathbf{B}^{(\ell)}_{i_k j_k}$ by the weighted maximum matching algorithm.
				
			\end{itemize}
			
			\State \textbf{4. } Compute the new sparse representations $\mathbf{X} = \mathcal{T}_s (2^{-\frac{q}{2}}(\mathbf{M}_q \cdots \mathbf{M}_1)^T \mathbf{Y})$.
		\end{algorithmic}
	\end{algorithm}
	
	B-transforms are numerically efficient structures, although they do involve multiplication operations. In this section we explore ways to reduce the numerical complexity even further. Consider the following structure:
	\begin{equation}
	\mathbf{M} = \prod_{\ell=1}^q \mathbf{\bar{M}}_\ell,\ \mathbf{\bar{M}}_\ell = \prod_{k=1}^{n/2} \mathbf{B}^{(\ell)}_{i_k j_k},
	\label{eq:theStransforms}
	\end{equation}
	with $\bigcup_{k=1}^{n/2} (i_k, j_k) = \{1,\dots, n\}  \text{ and } \bigcap_{k=1}^{n/2} (i_k, j_k)  = \emptyset$. This transformation is made up of $q$ stages. At each stage there are $\frac{n}{2}$ B-transforms that are chosen such that the coordinates $(i_k, j_k)$ are a partition of $\{1, \dots, n\}$. Consider now the objective function for a single block $\mathbf{\bar{M}}_1$
	\begin{equation}
	\| \mathbf{Y} - \mathbf{\bar{M}}_1 \mathbf{X} \|_F^2 = \| \mathbf{Y} \|_F^2 + \| \mathbf{X} \|_F^2 - 2\text{tr}(\mathbf{Z}) + \sum_{k=1}^{n/2} C_{i_k j_k}^{(t_k)},
	\label{eq:onlyoneM}
	\end{equation}
	such that the indices $(i_k, j_k)$ obey the constraints in \eqref{eq:theStransforms}. Optimizing the expression in \eqref{eq:onlyoneM}, i.e., minimizing the summation term by finding the best parings of the indices, is equivalent to the weighted maximum matching algorithm \cite{MMinG} (of maximum-cardinality matchings) on the graph with $n$ nodes and with edges $C_{i_k j_k}^{(0)} = \underset{t_k=1,\dots,8}{\min} - \ C_{i_k j_k}^{(t_k)}$ (the minus sign is set because we want to minimize the quantity). Because all $\mathbf{\bar{M}}_\ell$ are orthonormal, the manipulations in \eqref{eq:theBobjectivefunction} hold and therefore each $\mathbf{\bar{M}}_\ell$ can be updated while the others are fixed. We have found in our experimental settings that these iterative steps do not significantly improve the solution reached and therefore the algorithm builds the dictionary $\mathbf{M}$ in a single iteration, i.e., we set $K = 1$. This also highlights the importance of the initialization for $\mathbf{X}$, which is done again by the singular value decomposition.
	
	The full procedure is shown in Algorithm 3. The main difference with the previously introduced B$_m$--DLA is that at each step we update $\frac{n}{2}$ B-transforms simultaneously, not just one and none of these transforms use the same coordinates. In this fashion, the constant $2^{-\frac{1}{2}}$ factors out while keeping the $\mathbf{M}$ transformation orthonormal. 
	
	The transformation in \eqref{eq:theStransforms} can be equivalently written as $\mathbf{M} = 2^{-\frac{q}{2} } \prod_{\ell=1}^{q} \mathbf{M}_\ell$,
	where all $\mathbf{M}_\ell \in \mathbb{R}^{n \times n}$ are orthogonal sparse matrices with elements in $\{0, \pm 1 \}$, the diagonal only in $\{ \pm 1 \}$ and two non-zero entries per each row and column. Therefore, matrix-vector multiplication with one $\mathbf{M}_\ell$ takes $n$ additions and one multiplication and as such, matrix-vector multiplication with the whole $\mathbf{M}$-transform takes $nq$ additions and $n$ multiplications (or bit shifts if $q$ is even).	From this description of $\mathbf{M}$ it is easy to see that the coding complexity is approximately $\frac{q}{\ln 2}( n \ln n - n + 1)$ bits: the cost of encoding $q$ partitions of the indices, i.e., $q\sum_{i=1}^n \log_2 i = q \log_2 n! \approx \frac{q}{\ln 2}( n \ln n - n + 1)$ by Stirling's approximation. The constant $q$ is encoded implicitly as the number of partitions.
	
	The significant benefit of transforms like \eqref{eq:theStransforms} is that they avoid multiplications altogether. Furthermore, notice that each stage can be completely parallelized (since operations are done on distinct indices). The disadvantage, especially when compared to B$_m$--DLA, is that we force the transform to use all available indices and this constraint is sub-optimal in general and inferior to the choice \eqref{eq:Bargmin} made in B$_m$--DLA. Therefore, we expect M$_q$--DLA to perform worse than B$_m$--DLA in terms of the objective function value for the same number of basic transformations, i.e., $m = \frac{nq}{2}$.
	
	The computational complexity of M$_q$--DLA is dominated by the computation of $\mathbf{Z}$ which takes $O(n^2 N)$ operations and by the overall iterative process which takes a total of $O(qn^3)$ ($q$ times we have to perform the partitioning of the indices by the maximum matching algorithm).
	
	\noindent \textbf{Remark 3 (On the computational complexity of finding the best partition of indices).} The weighted maximum matching algorithm has complexity $O(n^3)$ which might be prohibitively large in some learning situations -- especially as compared to B$_m$--DLA which has a $O(n^2)$ complexity per iteration. An alternative is to use a sub-optimal, greedy, approach to build the indices partition. Consider a procedure that builds the partition in two iterative steps: compute the $\mathbf{B}^{(\ell)}_{i_k j_k}$ by $(i_k, j_k, t_k) = \underset{t,i<j;\ i,j \notin \mathcal{S}}{\arg \min} \ - 2\text{tr}(\mathbf{Z}_\ell) + C_{i j}^{(t)}$ and then update the set $\mathcal{S} \leftarrow \mathcal{S} \cup (i_k, j_k)$ for $k=1,\dots,\frac{n}{2}$ starting from $\mathcal{S} = \emptyset$. $\hfill \blacksquare$
	
	\noindent \textbf{Remark 4 (Another strategy for avoiding multiplication operations).} Notice that the matrices in \eqref{eq:localstructure1} are permutations with sign flips of the $2 \times 2$ Hadamard matrix $\frac{1}{\sqrt{2}} \begin{bmatrix}
	1 & 1\\ 1 & -1
	\end{bmatrix}$. We can extend these structures by using the $4 \times 4$ Hadamard matrix $\frac{1}{2} \left(  \begin{bmatrix}
	1 & 1\\ 1 & -1
	\end{bmatrix} \otimes \begin{bmatrix}
	1 & 1\\ 1 & -1
	\end{bmatrix} \right)$, whose scaling factor is now simple, i.e., a  power of two, and avoids multiplication operations. Unfortunately, operating on more than two coordinates increases the numerical complexity of the learning procedure, i.e., for each pair of four indices, instead of 8 options in \eqref{eq:localstructure1} we now have $768$ options, equivalent to all possible permutations of the four rows and columns and sign changes ($2 \times 2^4 \times  4!$) and there are ${n \choose 4} \approx \frac{n^4}{24}$ such pairs of indices, instead of $\frac{n(n-1)}{2}$ as for B$_m$--DLA. The total overall cost of one training iteration would therefore be dominated by the computation of the approximately $32n^4$ quantities $C_{i_k j_k r_k p_k}^{(t_k)}$ from which the minimum has to be found. Operating on even more coordinates simultaneously seems unreasonable from a computational perspective (in the learning phase).
	
	Still, the benefit is that the matrix-vector multiplication with a single such structure takes $12$ addition and $4$ bit shift operations. Therefore, an algorithm that uses these fundamental building blocks produces a transformation that completely avoids the multiplication operations. Finally, note that there are 5378 possibilities when considering all $3 \times 3$ orthonormal matrices (with different scaling factors $\sqrt{2}$, $\sqrt{3}$ and $2$) with entries in $\{0, \pm 1\}$, i.e., if we also allow zero entries -- these structure include \eqref{eq:localstructure1} and \eqref{eq:localstructure1b}. Therefore, this structure can be used only in scenarios where the learning time is not fundamentally constrained by time or power considerations.
	
	The advantage is that, as with the other algorithms described in this paper, parallelization is trivial. Furthermore, this approach would combine two of the major benefits of B$_m$--DLA and M$_q$--DLA: no calculations of partitions are necessary, i.e., we are not forced to repeatedly use coordinates that do not lead to significant reductions in the objective function, and there are no multiplication operations.
	
	We call this approach B$_m^\otimes$--DLA, and it follows the same steps as B$_m$--DLA but for only $K=1$ number of iterations. For brevity we omit the full description of the algorithm.$\hfill \blacksquare$
	
	Related to our previously stated desired computational properties we have that: P1 is achieved by taking $q \sim O(\log n)$ and since $\mathbf{M}^{-1} = \mathbf{M}^T$ we have the same computational benefits for the inverse transformation; P2, P3, and P4 are trivially achieved when $q$ is fixed to be even and therefore the algorithm avoids completely any multiplication operations.

	\section{The general case}
	
	In this section, we propose an algorithm to learn general dictionaries which have controllable complexity, including a variant which completely avoids multiplication operations. We begin by discussing the properties of scaling and shear transformations and then propose the learning procedure.
	
	\subsection{Shear transformations}
	
	Consider the set of shear $2 \times 2$ matrices
	\begin{equation}
	\mathcal{G}_3 = \left\{  \begin{bmatrix}
	1 & 0 \\
	b & 1
	\end{bmatrix},\ \begin{bmatrix}
	1 & c \\
	0 & 1
	\end{bmatrix}  \right\},\ b,c \in \mathbb{R},
	\label{eq:thetwoshears}
	\end{equation}
	and define a shear transformation $\mathbf{S}_{ij} \in \mathbb{R}^{n \times n}$ which is achieved for \eqref{eq:Rtransform} when $\mathbf{\tilde{R}}_{ij} \in \mathcal{G}_3$, i.e., $a = 1$ and $d=1$ fixed while $b$ and $c$ are free parameters or set to zero, alternatively.

	The objective function of our learning problem when the dictionary is a single $\mathbf{S}_{ij}$ now leads to
	\begin{equation}
	\min \ \| \mathbf{Y} - \mathbf{S}_{ij} \mathbf{X} \|_F^2 = \| \mathbf{Y} \|_F^2 + \| \mathbf{X} \|_F^2 - 2\text{tr}(\mathbf{Z}) + D_{ij}^{(t)},
	\label{eq:onlyoneS}
	\end{equation}
	for $t \in \{1, 2\}$, where $D_{ij}^{(1)} =b^2W_{ii} + 2b(W_{ij} - Z_{ji})$ and $D_{ij}^{(2)} = c^2W_{jj} + 2c(W_{ji} - Z_{ij})$. The minima, for $i=1,\dots,n$ and $j=i+1,\dots, n$, are
	\begin{equation}
	D_{ij}^{(1)} \! =\!  -(Z_{ji} - W_{ij})^2 W_{ii}^{-1},D_{ij}^{(2)} \! = \!  -(Z_{ij} - W_{ji})^2 W_{jj}^{-1},
	\label{eq:theDi_init}
	\end{equation}
	and are achieved for the optimum choices
	\begin{equation}
	b^\star = (Z_{ji} - W_{ij})W_{ii}^{-1} \text{ and } c^\star = (Z_{ij} - W_{ji})W_{jj}^{-1},
	\label{eq:optimalbc}
	\end{equation}
	respectively. Starting with all transformations $\mathbf{S}_{i_k j_k}$ set to the identity matrix, each one of the transformations $\mathbf{S}_{i_k j_k}$ is initialized sequentially in this fashion for $k = 1,\dots, m$.
	
	\noindent \textbf{Remark 5 (Optimality condition).} A necessary and sufficient condition for local optimality is that $D_{ij}^{(1)} = D_{ij}^{(2)} = 0$ and therefore $\mathbf{x}_j^T (\mathbf{y}_i - \mathbf{x}_i) = 0,\ \forall \ i\neq j$, i.e., in the spirit of the least squares solution applied row-wise, we have that any error row $\mathbf{\epsilon}_i = \mathbf{y}_i - \mathbf{x}_i$ is orthogonal to all rows $\mathbf{x}_i^T$ of $\mathbf{X}$.$\hfill \blacksquare$
	
	Now, after the initialization process, each $\mathbf{S}_{i_k j_k}$ is updated again while all other S-transforms are kept fixed. The objective function develops now to
	\begin{equation}
	\begin{aligned}
	\| &\mathbf{Y} \! - \! \mathbf{A}_k \mathbf{S}_{i_k j_k} \mathbf{X}_k \|_F^2 \! = \! \| \mathbf{y} \! - \! (\mathbf{X}_k^T \otimes \mathbf{A}_k) \text{vec}(\mathbf{S}_{i_k j_k}) \|_F^2 \\
	= & \| \mathbf{y} - (\mathbf{X}_k^T \otimes \mathbf{A}_k)\text{vec}(\mathbf{I}) - (\mathbf{X}_k^T \otimes \mathbf{A}_k) \text{vec}(\mathbf{L}_{i_k j_k}) \|_F^2 \\
	= & \| \mathbf{f}_k - (\mathbf{X}_k^T \otimes \mathbf{A}_k) \text{vec}(\mathbf{L}_{i_k j_k})  \|_F^2,
	\end{aligned}
	\label{eq:messydevelopment}
	\end{equation}
	where $\mathbf{y} = \text{vec}(\mathbf{Y})$, $\mathbf{B}_k = \mathbf{X}_k^T \odot \mathbf{A}_k \in \mathbb{R}^{nN \times n}$, $\odot$ is the Khatri-Rao product, $\mathbf{f}_k = \mathbf{y} - \mathbf{B}_k \mathbf{1}_{n \times 1}$, $\mathbf{A}_k = \prod_{q=k+1}^{m} \mathbf{S}_{i_q j_q}$, $\mathbf{X}_k = \prod_{q=1}^{k-1} \mathbf{S}_{i_q j_q} \mathbf{X}$, $\mathbf{L}_{i_k j_k} \in \{ b_k \mathbf{e}_{j_k}\mathbf{e}_{i_k}^T, c_k \mathbf{e}_{i_k} \mathbf{e}_{j_k}^T  \}$ and $\{ \mathbf{e}_i \}_{i=1}^n$ are the standard basis vectors of $\mathbb{R}^n$. We have used that $\text{vec}(\mathbf{ABC}) = (\mathbf{C}^T \otimes \mathbf{A} )\text{vec}(\mathbf{B})$. Notice that $(\mathbf{X}_k^T \otimes \mathbf{A}_k) \text{vec}(\mathbf{e}_{j_k} \mathbf{e}_{i_k}^T)$ selects the $(j_k+(i_k-1)n)^\text{th}$ column, i.e., $\mathbf{x}_{i_k} \otimes \mathbf{a}_{j_k}$,  while $(\mathbf{X}_k^T \otimes \mathbf{A}_k) \text{vec}(\mathbf{e}_{i_k} \mathbf{e}_{j_k}^T)$ selects the $(i_k+(j_k-1)n)^\text{th}$ column, i.e., $\mathbf{x}_{j_k} \otimes \mathbf{a}_{i_k}$. 
	
	To minimize the quantity in \eqref{eq:messydevelopment} the optimal choices are 
	\begin{equation}
	b^\star_k = \frac{  \mathbf{f}_k^T (\mathbf{x}_{i_k} \otimes \mathbf{a}_{j_k}) }{\| \mathbf{x}_{i_k} \|_2^2 \| \mathbf{a}_{j_k} \|_2^2 } \text{ and } c^\star_k = \frac{ \mathbf{f}_k^T (\mathbf{x}_{j_k} \otimes \mathbf{a}_{i_k}) }{\| \mathbf{x}_{j_k} \|_2^2 \| \mathbf{a}_{i_k} \|_2^2 },
	\label{eq:optimalbc2}
	\end{equation}
	respectively, and the minimum objective function values are
	\begin{equation}
	\min \ \| \mathbf{Y} - \mathbf{A}_k \mathbf{S}_{i_k j_k} \mathbf{X}_k \|_F^2 = \| \mathbf{f}_k \|_2^2 - E_{ij}^{(t)}, \text{ with}
	\label{eq:onlyoneSinthemiddle}
	\end{equation}
	\begin{equation}
	E_{ij}^{(1)} =
	\frac{ ( \mathbf{f}_k^T(\mathbf{x}_{i_k} \otimes \mathbf{a}_{j_k})  )^2 }{\| \mathbf{x}_{i_k} \|_2^2 \| \mathbf{a}_{j_k} \|_2^2 },
	\ E_{ij}^{(2)} = \frac{ ( \mathbf{f}_k^T (\mathbf{x}_{j_k} \otimes \mathbf{a}_{i_k}) )^2 }{\| \mathbf{x}_{j_k} \|_2^2 \| \mathbf{a}_{i_k} \|_2^2 },
	\label{eq:theonlyE}
	\end{equation}
	for all $i=1,\dots,n$ and $j=i+1,\dots,n$.
	
	As with the previously introduced structures, shear transformations have good numerical properties, i.e., matrix-vector multiplication $\mathbf{S}_{i_k j_k} \mathbf{x}$ takes one addition and one multiplication operation. If the coefficients $b$ or $c$ are represented in $\mathcal{R}_p$ then $\mathbf{S}_{i_k j_k, p} \mathbf{x}$ takes $p$ bit shifts and $p$ additions. Furthermore, inverses $\mathbf{S}_{i_k j_k}^{-1}$ shears themselves and easy to compute because $\begin{bmatrix}
	1 & 0 \\
	b & 1
	\end{bmatrix}^{-1} = \begin{bmatrix}
	1 & 0 \\
	-b & 1
	\end{bmatrix}$ and $\begin{bmatrix}
	1 & c \\
	0 & 1
	\end{bmatrix}^{-1} = \begin{bmatrix}
	1 & -c \\
	0 & 1
	\end{bmatrix}$
	and they have the same numerical properties as the direct shear transformations. There is no scaling factor for these inverses since $\det(\mathbf{S}_{i_k  j_k}) = 1$ always.
	
	To encode one shear transformation we need approximately $1 + C + 2\log_2 n$ bits (one bit to encode the choice between the two shears in \eqref{eq:thetwoshears}, the constant $C$ is the cost of encoding of $b$ or $c$, say $C=64$ for a double float, while the second term encodes the indices $i$ and $j$). 
	
	\subsection{Scaling transformations}
	\label{sec:scalings}
	
	Consider the R-transform in \eqref{eq:Rtransform} constrained to $b = 0, c = 0$, $d = 1$ and we drop the index $j$ which is now unnecessary to obtain a scaling matrix along a single coordinate
	\begin{equation}
	\mathbf{S}_i = \text{diag}(\begin{bmatrix}
	\mathbf{1}_{(i-1) \times 1} & a & \mathbf{1}_{(n-i) \times 1}
	\end{bmatrix}),\ a \in \mathbb{R}.
	\label{eq:theScaling}
	\end{equation}
	
	Similarly to \eqref{eq:onlyoneS}, with this scaling as the dictionary, the objective function of our learning problem is now
	\begin{equation}
	\min\ \| \mathbf{Y} - \mathbf{S}_i \mathbf{X} \|_F^2 = \|\mathbf{Y}\|_F^2 + \| \mathbf{X} \|_F^2 - 2\text{tr}(\mathbf{Z}) + F_i,
	\label{eq:theF}
	\end{equation}
	where we have denoted $F_i =  - 2Z_{ii}(a^\star - 1) + W_{ii}((a^\star)^2 -1)$ and used the scalar least squares solution $a^\star = Z_{ii} W_{ii}^{-1}$ that minimizes $\|\mathbf{y}_i - a^\star \mathbf{x}_i  \|_F^2$. Our goal is to find $\alpha$ such that $| \alpha a^\star | = 2^\delta,\ \delta \in \mathbb{Z},$ and $\| \mathbf{y}_i - \alpha a^\star \mathbf{x}_i \|_F^2$ is minimized. It is necessary to verify that the further scaling $\alpha$ is such that $\| \mathbf{y}_i - \mathbf{x}_i \|_F^2 - \| \mathbf{y}_i - \alpha a^\star \mathbf{x}_i \|_F^2 \geq 0$, which is obeyed when $- \left| \frac{1- a^\star}{a^\star} \right| \leq \alpha - 1 \leq \left|\frac{1-a^\star}{a^\star} \right|$, i.e., our scaling does not increase the objective function as compared to doing nothing along the $i^\text{th}$ coordinate. The left-hand side of the previous inequality is maximized when $\alpha = 2^\delta| a^\star |^{-1}$ is closest to one and therefore
	\begin{equation}
	\alpha^\star = 2^{[\log_2 | a^\star |]} | a^\star |^{-1} \text{ for }  a^\star = Z_{ii} W_{ii}^{-1}.
	\label{eq:alphastar}
	\end{equation}
	As such, the minimizer of \eqref{eq:theF} that is constrained to be a power of two and therefore its objective function value is
	\begin{equation}
	F_i = - 2Z_{ii}(\alpha^\star a^\star - 1) + W_{ii}((\alpha^\star a^\star)^2 -1).
	\label{eq:theFi_init}
	\end{equation}
	
	Consider now a scenario where each scaling transform was initialized and each $\mathbf{S}_{i_k}$ is updated again to further reduce the objective function while all others are kept fixed. Similarly to \eqref{eq:messydevelopment}, the objective function develops now to
	\begin{equation}
	\begin{aligned}
	\| \! \mathbf{Y} & - \! \mathbf{A}_k \mathbf{S}_{i_k} \mathbf{X}_k \|_F^2 \! = \!  \| \text{vec}(\mathbf{Y}) \! - \! (\mathbf{X}_k^T \! \otimes \! \mathbf{A}_k ) \text{vec}(\mathbf{S}_{i_k}) \|_F^2 \\
	= & \| \mathbf{y} - \mathbf{B}_k \mathbf{1}_{n \times 1} - (a_k - 1)\mathbf{B}_{k} \mathbf{e}_{i_k} \|_F^2 \\
	= & \| \mathbf{f}_k - (a_k - 1)\mathbf{B}_k\mathbf{e}_{i_k} \|_F^2,
	\end{aligned}
	\label{eq:scalingdevelopment}
	\end{equation}
	where $\mathbf{A}_k = \prod_{q=k+1}^{m} \mathbf{S}_{i_q}$, $\mathbf{X}_k = \prod_{q=1}^{k-1} \mathbf{S}_{i_q} \mathbf{X}$ and the structure of $\mathbf{S}_{i_k} = \mathbf{I}_{n \times n} + (a_k -1)\text{diag}(\mathbf{e}_{i_k})$. For the other variables we have used here the same notation as in \eqref{eq:messydevelopment}. Since we want to minimize the quantity in \eqref{eq:scalingdevelopment}, then we have
	\begin{equation}
	\min\ \| \mathbf{Y} - \mathbf{A}_k \mathbf{S}_{i_k} \mathbf{X}_k \|_F^2 =  \| \mathbf{f}_k \|_2^2 - G_i,
	\end{equation}
	and using the minimizer of this expression
	\begin{equation}
	a_k^\star = 
	\frac{\mathbf{f}_k^T \mathbf{b}_{i_k} }{\| \mathbf{b}_{i_k} \|_2^2} +1 = \frac{\mathbf{f}_k^T (\mathbf{x}_{i_k} \otimes \mathbf{a}_{i_k})}{\| \mathbf{x}_{i_k} \|_2^2 \| \mathbf{a}_{i_k} \|_2^2} + 1,
	\label{eq:optimalgamma}
	\end{equation}
	it follows that for $i=1,\dots,n$ we have
	\begin{equation}
	G_i = \frac{(\mathbf{f}_k^T \mathbf{b}_{i_k})^2}{\| \mathbf{b}_{i_k} \|_2^2} = \frac{(\mathbf{f}_k^T (\mathbf{x}_{i_k} \otimes \mathbf{a}_{i_k}))^2}{\| \mathbf{x}_{i_k} \|_2^2 \| \mathbf{a}_{i_k} \|_2^2},
	\label{eq:theoneG}
	\end{equation}
	where $\mathbf{b}_{i_k} = \mathbf{x}_{i_k} \otimes \mathbf{a}_{i_k}$ is the $i_k^\text{th}$ column of $\mathbf{B}_k$ and $\mathbf{x}_{i_k}^T$ is the $i_k^\text{th}$ row of $\mathbf{X}_k$ and $\mathbf{a}_{i_k}$ is the $i_k^\text{th}$ column of $\mathbf{A}_k$. We have used the fact that the squared $\ell_2$ norm of a Kronecker product is the product of the squared $\ell_2$ norms of the two vectors involved.
	
	Due to their simplicity, the scaling transformations $\mathbf{S}_i$ are numerically efficient, i.e., matrix-vector multiplication $\mathbf{S}_i \mathbf{x}$ takes only one bit shift or one multiplication operation (if we set $\alpha^\star = 1$ and just solve a general unconstrained least squares problem on the $i^\text{th}$ coordinate). The inverse $\mathbf{S}_{i}^{-1}$ is also a scaling transformation (with $a^{-1}$ on position $i$) with the same numerical properties as the direct transformation.
	
	To encode one scaling transformation we need $C + \log_2 n$ bits (the constant is the cost of encoding of $a$, say $C=64$ for a double float, while the second term encodes the index $i$). Notice that we could represent $a$ in $\mathcal{R}_p$, for a given $p$, but then $a^{-1}$ does not have, in general, a representation in $\mathcal{R}_{p'}$ for any finite $p'$. It is for this reason that when it comes to the scaling transformation and its scaling factor $a$ we allow only two possibilities: either we take $a^\star$ the optimum least squares choice if we are working with arbitrary precision, i.e., $p = \infty$, or we take $\alpha^\star a^\star$ to ensure the scaling is a bit shift and therefore ensure the consistency of the inverse scaling transformation.
	
	\begin{algorithm}[t]
		\caption{ \textbf{-- S$_{m,p}$--DLA. } \newline \textbf{Input: } The dataset $\mathbf{Y} \in \mathbb{R}^{n \times N}$, the sparsity $s$, the precision $p$, the number of scalings and shears $m$ in the dictionary.\newline \textbf{Output: } The general transformation $\mathbf{S}$ \eqref{eq:theSStransforms} composed of $m$ scalings and shears, and the sparse representations $\mathbf{X}$ such that $\| \mathbf{Y} - \mathbf{SX} \|_F^2$ is reduced.}
		\begin{algorithmic}
			\State \textbf{1. } Initialize transform: set $\mathbf{S}_{i_k j_k} = \mathbf{I}_{n \times n}$ for $k=1,\dots,m$.
			
			\State \textbf{2. } Initialize sparse representations: compute the singular value decomposition of the dataset $\mathbf{Y} = \mathbf{U \Sigma V}^T$ and compute the sparse representations $\mathbf{X} = \mathcal{T}_s (\mathbf{U}^T \mathbf{Y})$.
			
			\State \textbf{3. } Initialization of transformations:
			\begin{itemize}
				
				\item Compute $\mathbf{Z}$ and $\mathbf{W}$ by \eqref{eq:theZ} and \eqref{eq:theW}, respectively.
				
				\item Use \eqref{eq:theFi_init} and \eqref{eq:theDi_init} compute all the scores $D_{ij}^{(t)}$ and $F_i$, $t \in \{1,2\}$, $i=1,\dots, n$ and $j = i+1,\dots,n$.
				
				\item For $k = 1,\dots,m:$
				\begin{itemize}
					\item Initialize $\mathbf{S}_{i_k j_k}$ by searching for the minimum $\{ D_{ij}^{(t)}, F_i \}$ across all indices and compute the optimal transformation values: $\alpha^\star a^\star$ with \eqref{eq:alphastar} if $p \neq \infty$ or $\alpha^\star = 1$ and $a^\star$ by \eqref{eq:alphastar} otherwise, and by representing $b^\star$ or $c^\star$ \eqref{eq:optimalbc} in the set $\mathcal{R}_p$.
					
					\item Update scores: $D_{i j_k}^{(t)}$, $D_{j_k j}^{(t)}$, $D_{i_k j}^{(t)}$, $D_{i i_k}^{(t)}$ for $i=1,\dots, n$ and $j = i+1,\dots,n$, $F_{i_k}$ and $F_{j_k}$, $t \in \{1,2\}$.
				\end{itemize}
				
				\item Compute sparse representations $\mathbf{X} = \text{OMP}(\mathbf{Y}, \mathbf{S}, s)$.
			\end{itemize}
			\State \textbf{4. } For $1,\dots,K$:
			\begin{itemize}
				\item Update each transformation in the factorization:
				
				\begin{itemize}
					
					\item Compute all the scores $E_{ij}^{(t)}$ and $G_i$, $t \in \{1,2\}$, $i=1,\dots, n$ and $j = i+1,\dots,n$.

					\item With all $\mathbf{S}_{i_q j_q},\ q \neq k,$ fixed, compute the new $\mathbf{S}_{i_k j_k}$ the minimizer of \eqref{eq:theoneG} or \eqref{eq:theonlyE} by searching for the minimum $\{ E_{ij}^{(t)},  G_i \}$ across all indices and compute the optimal transformation values: $\alpha^\star \gamma^\star$ with \eqref{eq:alphastar} and \eqref{eq:optimalgamma} if $p \neq \infty$ or $\alpha^\star = 1$ and $a^\star$ by \eqref{eq:optimalgamma} otherwise, and by representing $b^\star$ or $c^\star$ \eqref{eq:optimalbc2} in the set $\mathcal{R}_p$.
				\end{itemize}

				\item Compute sparse representations $\mathbf{X} = \text{OMP}(\mathbf{Y}, \mathbf{S}, s)$.
			\end{itemize}
		\end{algorithmic}
	\end{algorithm}
	\subsection{Numerically efficient general transforms: S$_{m,p}$--DLA}
	
	Similarly to the other transformations described in this paper, we consider now the following dictionary structure
	\begin{equation}
	\mathbf{S} =  \prod_{k=1}^{m} \mathbf{S}_{i_k j_k}= \mathbf{S}_{i_m j_m} \dots \mathbf{S}_{i_1 j_1},
	\label{eq:theSStransforms}
	\end{equation}
	where each $\mathbf{S}_{i_k j_k}$ is either a scaling or a shear transformation. For convenience, we denote $m_1$ and $m_2$ the number of scalings and shears, respectively -- we have $m_1+m_2 = m$.

	The complete learning method is described in Algorithm 4. This procedure has two main components: the initialization phase and the iterative process that improves the factorization.
	
	In the initialization phase, we use again the singular value decomposition, assume that our initial dictionary is $\mathbf{U}$ and then proceed to compute the first sparse representations $\mathbf{X}$. Then we proceed to initialize each $\mathbf{S}_{i_k j_k}$ iteratively. The computational cost of this step is dominated by the calculation of all the scores which takes $O(n^3 N)$: there are $n^2$ scores and we need $O(nN)$ to compute each one (it is the computational cost of performing the dot product between $\mathbf{f}_k$ and the Kronecker product). Then, because we are dealing with a general dictionary, we use the batch Orthogonal Matching Pursuit (OMP) algorithm \cite{OMP, AKSVD} to build $\mathbf{X}$.
	
	The iterative process now tries to improve the accuracy of the factorization by updating each individual transformation while all others are kept fixed. When we update a transformation we calculate the indices $(i_k, j_k)$ and the coefficients of the transformation $a, b$ or $c$. For this reason, this iterative step is computationally expensive at takes $O(n^3mNK)$: there are $K$ iterations which update each of the $m$ transformations and for each we need to compute $n^2$ scores where the computing load for a single one is dominated a dot product between $\mathbf{f}_k$ and a Kronecker product which takes $O(nN)$. Luckily, in practice we observer that a low number of iterations will usually suffice to reach a very low representation error $\| \mathbf{Y} - \mathbf{SX} \|_F^2$. After all the transformations are updated, we also update the representations $\mathbf{X}$ again by the OMP algorithm.
	
	We allow an input parameter $p$ to set the precision of the transformations we compute. In the case of the shear transforms, we will represent the coefficients $b$ and $c$ in the set $\mathcal{R}_p$ while for the scaling transformations we allow two options: either use the full precision $a$ computed for $p=\infty$ or approximate it by the nearest power of two whenever $p \neq \infty$. We treat the scaling transformation in this binary fashion in order to keep consistent the inverse scaling operation (see the discussion in Section \ref{sec:scalings}).

	Iteratively, we update each $\mathbf{S}_{i_k j_k}$ component of $\mathbf{S}$ and the sparse representations $\mathbf{X}$ via the OMP algorithm. Although the optimization steps to update the dictionary components are solved exactly and therefore always reduce the objective function, the OMP step cannot be guaranteed to reach optimal representations in general and therefore the algorithm is generally not guaranteed to converge monotonically to a local minimum point. As such, we track the best solution obtained so far in the iterative process and return it.
	
	Transforms built by S$_{m,\infty}$--DLA will perform $m$ multiplications and $m_2$ additions and those built by S$_{m,p}$--DLA will perform $m_1+ pm_2$ bit shifts and $pm_2$ additions when $p \neq \infty$. Therefore, the computational complexity of these transforms is not predetermined only by the choice of $m$ but also by the actual factorization which is constructed. Related to our previously stated desired computational properties we have that: P1 is achieved by taking $m \sim O(\log n)$ and since inverses of scalings and shears are themselves scalings and shears (this covers also P3) we have the same computational benefits for the inverse transformation; for P2 and P4 we need the scalings to perform shift operations and the shears to have $p < \infty$.
	
	\noindent \textbf{Remark 6 (Completeness of scalings and shears).} Every invertible $\mathbf{S}$ can be represented as a product of $n^2 - n$ shear and $n$ scaling transformations (note that this was not the case for the binary orthogonal building blocks, see Remark 2). The proof of this fact is constructive: perform Gaussian elimination on $\mathbf{S}$ to diagonalize it using $n^2 - n$ shears and then represent the resulting diagonal with $n$ scalings (permutations are also allowed if elimination is done with pivoting). Of course, in our approach we are trying to build $\mathbf{S}$ such that $m \ll n^2$, otherwise there is no computational benefit. $\hfill \blacksquare$
	
		\subsection{A note on the general overcomplete case}
		
		Throughout this paper we have considered learning square linear transformations. Many of the successful applications of dictionary learning involves overcomplete transformations, i.e., learning an $n \times d$ transformation with $d > n$. All building blocks that we have introduced are naturally square, e.g., the factorization \eqref{eq:theSStransforms}, but they can be made overcomplete by pre-multiplication with a mask matrix $\mathcal{M} = \begin{bmatrix}
		\mathbf{I}_{n \times n} & \mathbf{0}_{n \times (d-n)}
		\end{bmatrix}$. From an optimization perspective we are therefore in the situation of the quantities \eqref{eq:messydevelopment} and \eqref{eq:scalingdevelopment} where we replace $\mathbf{A}_k$ by $\mathcal{M} \mathbf{A}_k$. The approach to improve each building block is essentially the same with the notable exception that many scores are null.
	\begin{figure}[t]
		\centering
		\includegraphics[trim = 18 5 30 10, clip, width=0.33\textwidth]{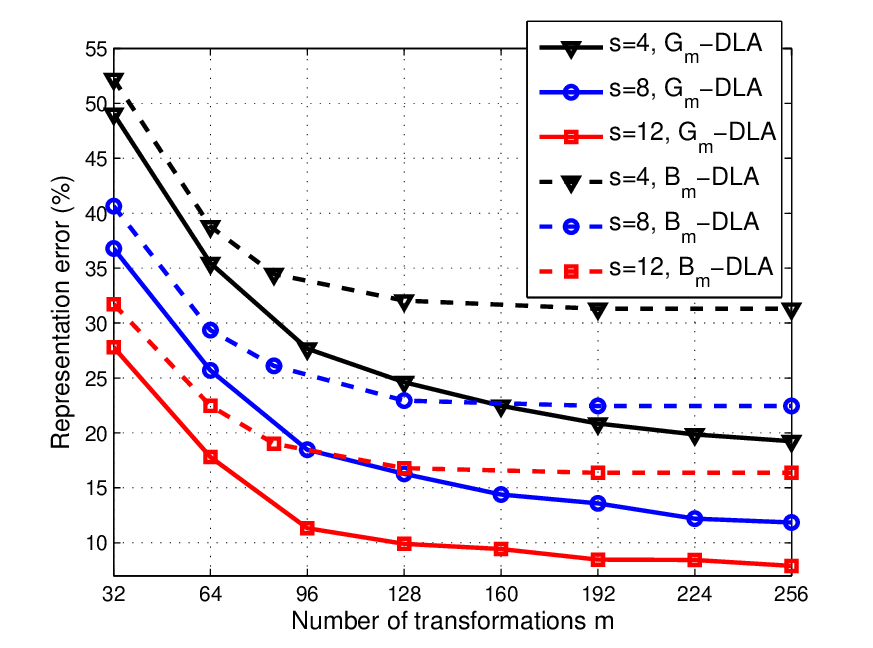}
		\caption{A comparison between G$_m$--DLA \cite{FastSparsifyingTransforms} and the proposed B$_m$--DLA using the representation error \eqref{eq:epsilon} for various sparsity levels $s$.}
		\label{fig:figure1}
	\end{figure}
	\begin{figure}[t]
		\centering
		\includegraphics[trim = 18 5 30 15, clip, width=0.33\textwidth]{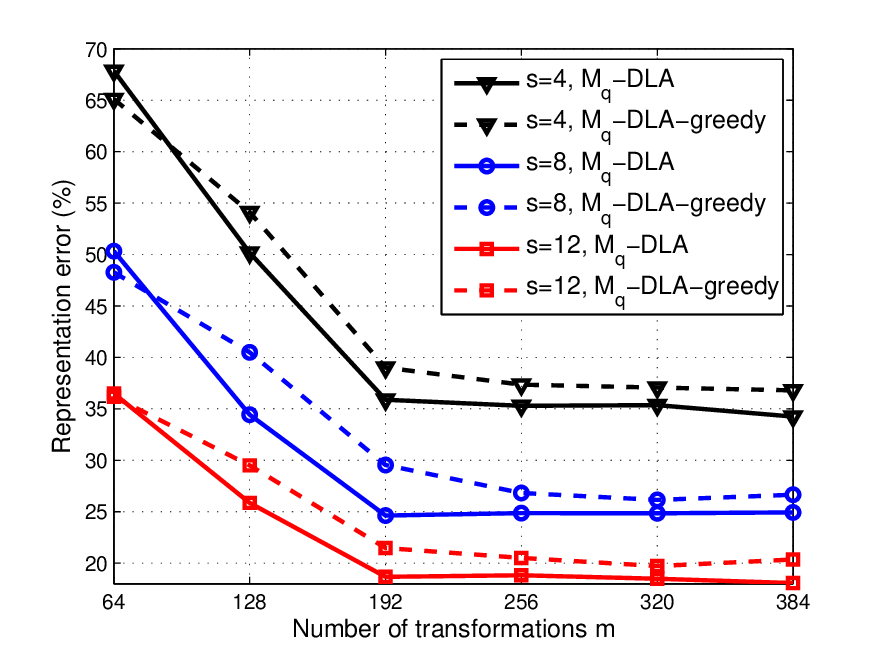}
		\caption{Representation errors \eqref{eq:epsilon} achieved by M$_q$--DLA and a variant called M$_q$--DLA-greedy which has lower training complexity (see Remark 3).}
		\label{fig:figure2}
	\end{figure}
	\section{Experimental results}
	
	In this section we test the proposed learning algorithms on image data, where we have well-known numerically efficient transformations against which to compare. The training data we consider is built from popular test images from the image processing literature (lena, peppers, boat etc.). The dataset $\mathbf{Y} \in \mathbb{R}^{64 \times 12288}$ consists of $8 \times 8$ non-overlapping image patches with their means removed. To evaluate the learning algorithms, we take the relative representation error of the dataset $\mathbf{Y}$ in the dictionary $\mathbf{D}$ with the representations $\mathbf{X}$ as
	\begin{equation}
	\epsilon = \| \mathbf{Y} - \mathbf{DX} \|_F^2 \| \mathbf{Y} \|_F^{-2}\ (\%).
	\label{eq:epsilon}
	\end{equation}
	
	\begin{figure}[t]
		\centering
		\includegraphics[trim = 18 5 30 10, clip, width=0.33\textwidth]{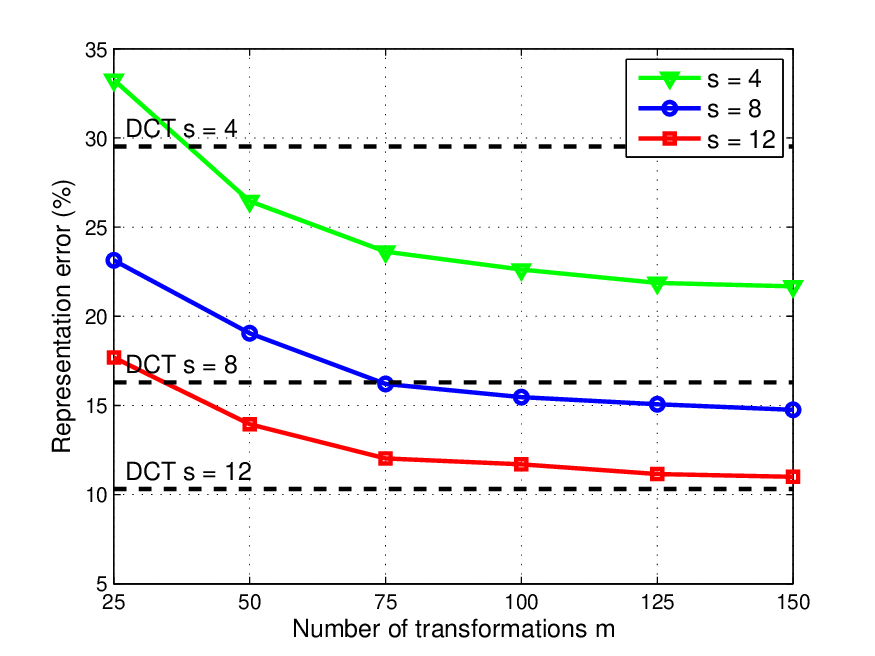}
		\caption{Representation errors achieved by B$_m^\otimes$--DLA (see Remark 4) for various sparsity levels $s$. For reference we show the DCT. B$_m^\otimes$--DLA reaches the computational complexity of the DCT for $m = \frac{1}{8}n \log_2 n = 48$.}
		\label{fig:figure3}
	\end{figure}
	\begin{figure}[t]
		\centering
		\includegraphics[trim = 18 5 30 15, clip, width=0.33\textwidth]{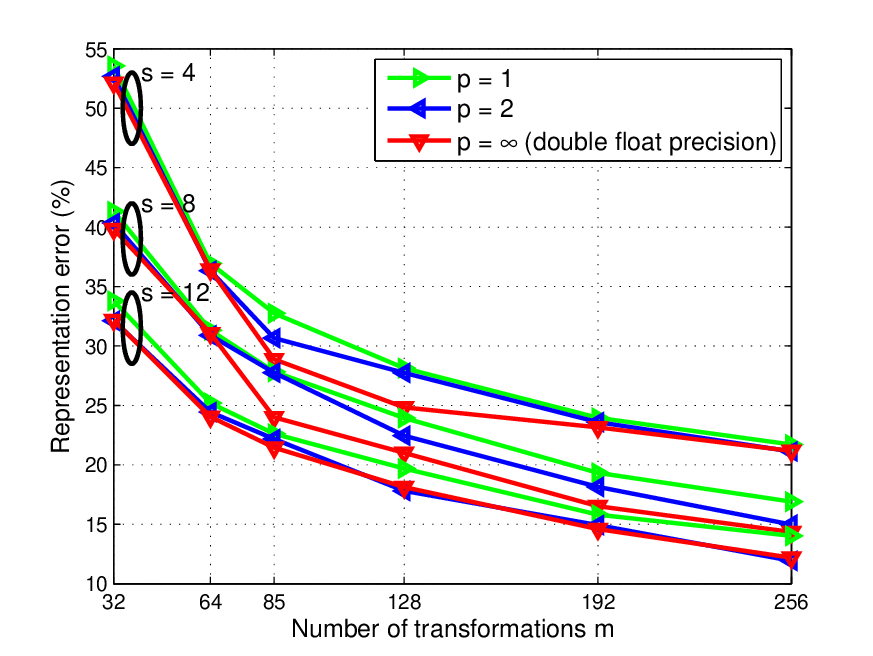}
		\caption{Representation errors achieved by S$_{m,p}$--DLA for various number of transformations $m$, sparsity levels $s$ and precision $p$.}
		\label{fig:figure4}
	\end{figure}
	The dictionary learning problem constraints require that the transformation we learn have $\ell_2$ normalized columns. This constraint is trivially obeyed for the orthonormal transforms designed via B$_m$--DLA and M$_q$--DLA but not in the case of those built by S$_{m,p}$--DLA. One solution is to introduce a diagonal matrix $\mathbf{D} \in \mathbb{R}^{n \times n}$ and update \eqref{eq:theStransforms} via $\mathbf{S} \leftarrow \mathbf{SD}$ such that columns are $\ell_2 $ normalized. Once the  representations are computed, $\mathbf{S}$ is restored to \eqref{eq:theSStransforms} and $\mathbf{X} \leftarrow \mathbf{DX}$.
	
	We compare against the FFT or the discrete cosine transform (DCT) which, for real-valued inputs, has complexity $2n\log_2 n$: $\frac{3}{2} n \log_2 n$ additions and $\frac{1}{2} n \log_2 n$ multiplications. We also compare against transforms built by G$_m$--DLA, which have complexity $6m$, even additions and multiplications.
	
	In Figure \ref{fig:figure1} we show the representation errors achieved by B$_m$--DLA and compare them against G$_m$--DLA \cite{FastSparsifyingTransforms}. We always expect G$_m$--DLA to perform better, as shown in the figure, especially for large $m$. Since up to $m = 64$ we have that B$_m$--DLA closely tracks G$_m$--DLA, this introduces the idea of a potential hybrid algorithm: use binary transformations in the beginning and then proceed with G$_m$--DLA when the decrease in the error slows or plateaus. The advantage of B$_m$--DLA transforms is that they are $33\%$ faster than G$_m$--DLA transforms. Similar results are shown in Figure \ref{fig:figure2} for M$_q$--DLA. The same plateau of the error is observed for $m \geq 128$ for all sparsity levels $s$. Because this approach is computationally expensive in the training phase, we also show a cheaper method called M$_q$--DLA (see Remark 3) that performs similarly. We complete the orthonormal learning experiments with Figure \ref{fig:figure3} where we show B$_m^\otimes$--DLA (see Remark 4) for $s \in \{ 4,8,12 \}$. Notice that the representation error drops faster than that of B$_m$--DLA and it plateaus for higher $m$. As explained, the disadvantage of B$_m^\otimes$--DLA is the increased training time: on a modern computer, calculating all scores $C_{i_k j_k r_k p_k}^{(t_k)}$ takes over three hours and updating the scores with each iteration is done only for a random subset to keep the running time for one iteration to only a few minutes. 
	
	In Figure \ref{fig:figure4} we show the representation errors achieved by S$_{m,p}$--DLA. The effect of the precision parameter $p$ is interesting. For any sparsity level $s$, when the number of transformations $m$ is low ($\leq 64$) the parameter $p$ does not play an important role and therefore $p=1$, which has the lowest computational complexity, is preferred. For other $m$, there are slight differences (mostly within $5\%$) in the error and, as expected, $p = \infty$ works best. For $s=4$ the proportion of scalings and shears in the transforms designed is approximately $10\%$ to $90\%$.
	
	\noindent \textbf{Remark 7 (Sub-optimal transform learning).} All proposed algorithms choose indices $i,j$ on which the linear transformation operate greedily (maximally reduce the current objective function). In some situations, it might be convenient to make sub-optimal choices that take into account other goals.
	
	For example, from a computational perspective, we might consider a highly local computational model, i.e., our algorithms perform operations, in place, on memory locations that are close such that they can exploit benefits of hierarchically memory structures (cache-oblivious algorithms \cite{Frigo:2012:CA:2071379.2071383}). In our case this might translate in constraints as $|i_k -j_k| \leq \epsilon$, i.e., for a particular transformation control the distance between the operands, and $|i_k - i_{k+1}| \leq \epsilon$ (and similarly for $j_k$, $j_{k+1}$), i.e., consecutive transformations operate on neighboring regions of memory. Another example is bounding the dynamic range \cite[Section IV]{IntegerFFT} of the intermediate stages of our transforms. For example, in the case of B$_m$--DLA, given an input $\mathbf{x}$ such that $|x_i| \leq C$ the output $\mathbf{y} = \mathbf{Bx}$ is such that $|y_i| \leq 2^{\frac{m}{2}} C$ (equality happens if the same operation takes places for the same indices $i,j$, e.g., $x_i \leftarrow 2^{-\frac{1}{2}} (x_i + x_j)$ happens $m$ times. Therefore, imposing some index diversity and/or making sure we do not have repetitive operations for the same indices will reduce the maximum possible output magnitude.$\hfill \blacksquare$
	
	\begin{figure}[t]
		\centering
		\includegraphics[trim = 18 5 28 10, clip, width=0.33\textwidth]{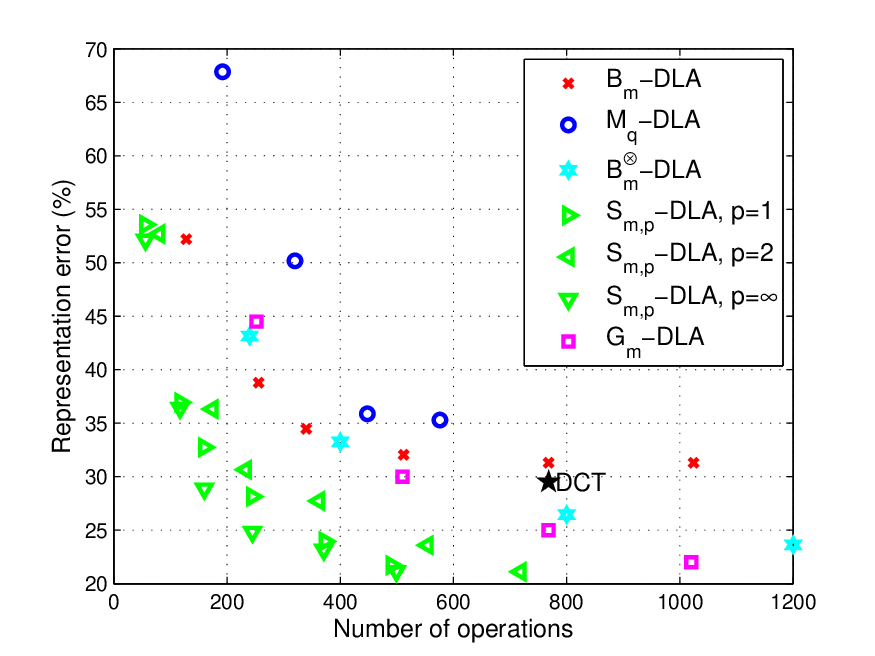}
		\caption{A comparative study of the transforms created by the proposed methods. We also show, for reference, the DCT and G$_m$--DLA \cite{FastSparsifyingTransforms}. We set $s = 4$. Number of operations counts everything for matrix-vector multiplication.}
		\label{fig:figure5}
	\end{figure}
	\begin{figure}[t]
		\centering
		\includegraphics[trim = 18 5 28 15, clip, width=0.33\textwidth]{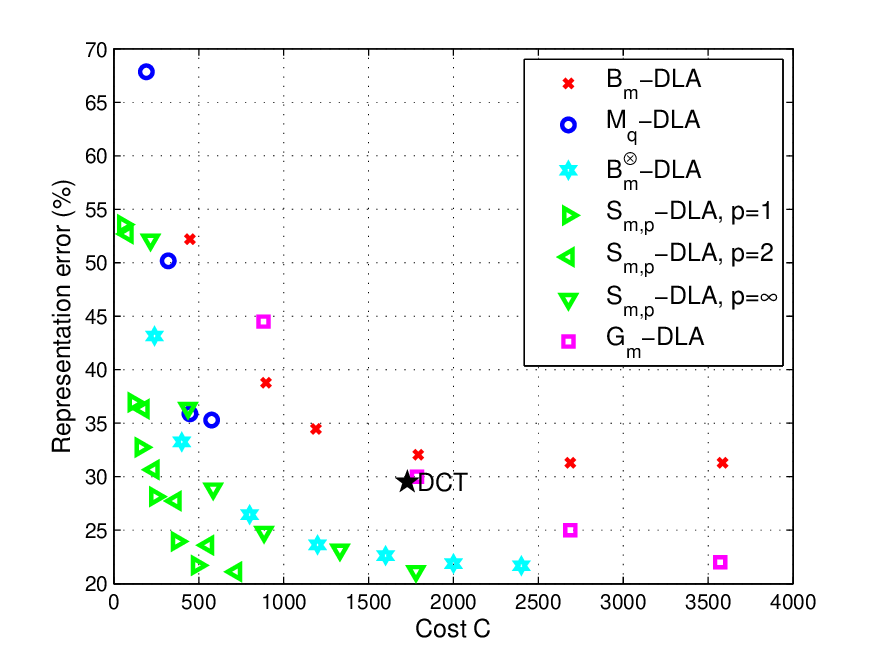}
		\caption{Similar to Figure \ref{fig:figure5} but we define a cost $C = A + \gamma M$, where $A$ is the number of additions and bit shifts, $M$ is the number of multiplications and $\gamma = 6$ is an extra penalty factor.}
		\label{fig:figure6}
	\end{figure}
	Finally, in Figures \ref{fig:figure5} and \ref{fig:figure6} we show a comparative study of the proposed transforms for fixed $s=4$. The point is to have a Pareto curve to show the computational-representation error trade-off: in the first plot we show the overall computational complexity and then, in the second plot, we apply an extra penalty $\gamma =6$ to multiplication operations. Unsurprisingly, the non-orthonormal transforms built by S$_{m,p}$--DLA are the most effective (for $p = \infty$ in Figure \ref{fig:figure5} and $p = 1$ in Figure \ref{fig:figure6}). The orthonormal transforms behave as expected as well: M$_q$--DLA performs very well in terms of computational complexity but worse in terms of representation errors (and plateaus quickly), B$_m$--DLA makes better progress and achieves better representation errors (although it also slows down progress with increased $m$) while B$_m^\otimes$--DLA combines the benefits of the two approaches with the drawback is that training took overnight. Note that, for the same complexity, B$_m$--DLA and B$_m^\otimes$--DLA are very close to the DCT while with the extra penalty $\gamma$ for multiplications B$_m^\otimes$--DLA outperforms the DCT, i.e., when addition and bit shift operations are cheap enough we can afford a large enough number of these operations to surmount the constraint of avoiding multiplications.

	\section{Conclusions}
	
	In this paper we have proposed several dictionary learning algorithms that produce linear transformation which exhibit low computational complexity in general and reduce (or completely eliminate) multiplications in particular. We show that these transforms perform very well on image data where we compare against the DCT. We show that multiplications can be avoided and representation errors of image data can be kept low by replacing multiplication operations with a larger number of addition and bit shift operations.
	
	\bibliographystyle{IEEEtran}
	\bibliography{refs}
	
\end{document}